\documentclass[11pt,a4paper]{article}
\usepackage{times,latexsym}
\usepackage{url}
\usepackage[T1]{fontenc}

   \usepackage[acceptedWithA]{tacl2021v1}
\usepackage{xspace,mfirstuc,tabulary}

\newif\iftaclinstructions
\taclinstructionsfalse 
\iftaclinstructions
\renewcommand{\confidential}{}
\renewcommand{\anonsubtext}{(No author info supplied here, for consistency with
TACL-submission anonymization requirements)}
\newcommand{\instr}
\fi

\iftaclpubformat 

\else

\fi

\usepackage[utf8]{inputenc}

\usepackage{microtype}

\usepackage{inconsolata}
\usepackage{times}
\usepackage{latexsym}
\usepackage{url}            %
\usepackage{wrapfig}
\usepackage{booktabs}       %

\usepackage{caption}
\usepackage{placeins}
\usepackage{amssymb}%
\usepackage{pifont}%
\usepackage{floatrow}
\usepackage{float}
\usepackage{graphicx}
\usepackage{fancyvrb}
\usepackage{amsmath}
\usepackage{amsfonts}       %
\usepackage{nicefrac}       %
\usepackage{xcolor}         %
\usepackage{xspace}         %
\usepackage{multirow}
\usepackage{array}
\usepackage{siunitx}
\usepackage{booktabs}
\usepackage{colortbl}
\usepackage{color}
\usepackage{soul}
\usepackage{tcolorbox}
\usepackage{hyperref}
\usepackage{enumitem}
\usepackage{amsmath}
\usepackage{amssymb}
\usepackage{tikz}
\usetikzlibrary{shapes,arrows,automata}
\usepackage{diagbox}
\usepackage{adjustbox}
\usepackage[edges]{forest}
\usepackage[11pt]{moresize}

\def\Snospace~{\S{}}

\newcommand\extrafootertext[1]{%
    \bgroup
    \renewcommand\thefootnote{\fnsymbol{footnote}}%
    \renewcommand\thempfootnote{\fnsymbol{mpfootnote}}%
    \footnotetext[0]{#1}%
    \egroup
}

\usepackage[T1]{fontenc}
\definecolor{mygray}{RGB}{220, 220, 220}
\definecolor{lightgreen}{RGB}{223,255,219}
\definecolor{lightred}{RGB}{255,219,219}

\definecolor{pos}{RGB}{167, 199, 231}
\definecolor{neg}{RGB}{250, 160, 160}

\usepackage{soul}

\newcommand\squad{SQuAD2\xspace}
\newcommand\qasper{QASPER\xspace}
\newcommand\pubmedqa{PubmedQA\xspace}

\newcolumntype{L}[1]{>{\raggedright\let\newline\\\arraybackslash\hspace{0pt}}p{#1}}
\newcolumntype{M}[1]{>{\raggedright\let\newline\\\arraybackslash\hspace{0pt}}m{#1}}
\newcolumntype{C}[1]{>{\centering\let\newline\\\arraybackslash\hspace{0pt}}m{#1}}
\newcolumntype{D}[1]{>{$\displaystyle}p{#1}<{$}}

\definecolor{darkgreen}{rgb}{0.0, 0.4, 0.13}

\newcolumntype{d}[1]{D{.}{.}{#1}}

\newcommand{\idea}{%
  \begingroup\normalfont
  \includegraphics[height=1.3\fontcharht\font`\B]{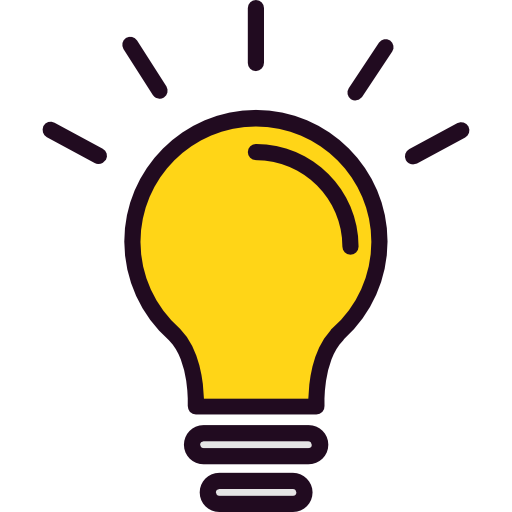}%
  \endgroup
}

\title{Know Your Limits: A Survey of Abstention in Large Language Models}

\author{Bingbing Wen\textsuperscript{1} \quad Jihan Yao\textsuperscript{1} \quad Shangbin Feng\textsuperscript{1} \quad
\textbf{Chenjun Xu\textsuperscript{1}} \\\textbf{Yulia Tsvetkov\textsuperscript{1}} \quad \textbf{Bill Howe\textsuperscript{1}} \quad \textbf{Lucy Lu Wang\textsuperscript{1,2}} \\
\textsuperscript{1}University of Washington \quad\textsuperscript{2}Allen Institute for AI \\
\small{\texttt{\{bingbw, chenjux, billhowe, lucylw\}@uw.edu}} \\
\small{\texttt{\{jihany2, shangbin, yuliats\}@cs.washington.edu}}
}

\date{}

\begin{document}
\maketitle

\begin{abstract}
\vspace{-2mm}
Abstention, the refusal of large language models (LLMs) to provide an answer, is increasingly recognized for its potential to mitigate hallucinations and enhance safety in LLM systems. In this survey, we introduce a framework to examine abstention from three perspectives: the query, the model, and human values. We organize the literature on abstention methods, benchmarks, and evaluation metrics using this framework, and discuss merits and limitations of prior work. We further identify and motivate areas for future research, such as whether abstention can be achieved as a meta-capability that transcends specific tasks or domains, and opportunities to optimize abstention abilities in specific contexts. In doing so, we aim to broaden the scope and impact of abstention methodologies in AI systems.\footnote{A list of abstention-related papers from this review can be found at \href{https://github.com/chenjux/abstention}{https://github.com/chenjux/abstention}.}
\end{abstract}

\vspace{-3mm}
\section{Introduction}
\vspace{-1mm}
Large language models (LLMs) have demonstrated generalization capabilities across NLP tasks such as question answering (QA)~\citep{wei2021finetuned, chowdhery2022palm}, abstractive summarization~\citep{Zhang2023BenchmarkingLL}, and dialogue generation~\citep{yi2024survey}.  But these models are also unreliable, having a tendency to ``hallucinate'' false information in their responses~\cite{ji2023survey}, 
generate overly certain or authoritative responses~\cite{zhou2024relying}, answer with incomplete information~\cite{zhou2023contextfaithful}, or produce harmful or dangerous responses~\cite{anwar2024foundational}. 
In these situations, the model should ideally \emph{abstain}: to refuse to answer in the face of uncertainty~\citep{wen-etal-2024-characterizing, feng2024don, yang2023alignment}. 

\begin{figure}
\centering
\includegraphics[width=\linewidth]{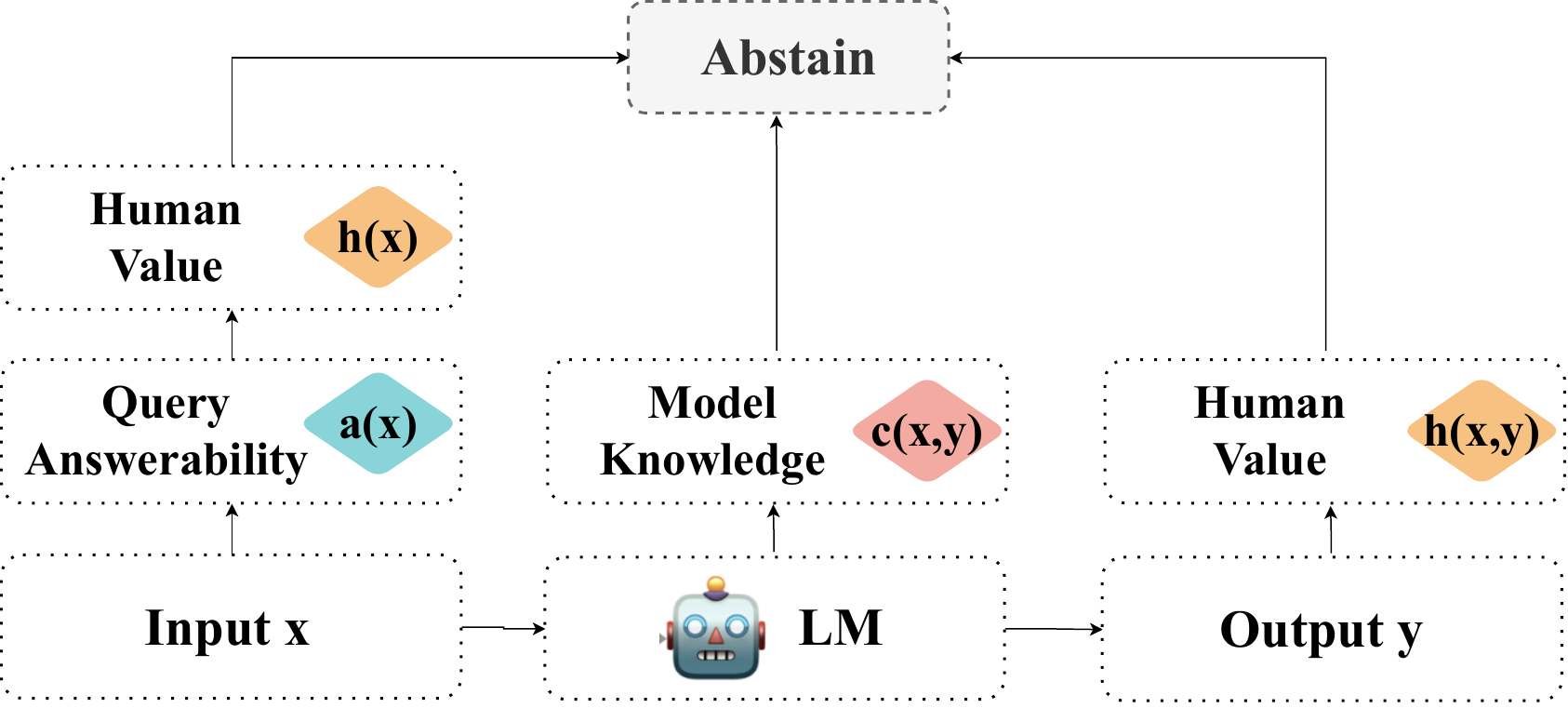}
\caption{Our proposed framework for abstention in language models. Starting with input query $\mathbf{x}$, the query can be gauged for answerability $a(\mathbf{x})$ and alignment with human values $h(\mathbf{x})$. 
The model then generates a potential response $\mathbf{y}$ based on the input $\mathbf{x}$. If query conditions are not met, the model's confidence in the response $c(\mathbf{x},\mathbf{y})$ is too low, or if the response's alignment with human values $h(\mathbf{x},\mathbf{y})$ is too low, the system should abstain. 
}\label{fig:system}
\vspace{-2mm}
\end{figure}

Current methods to encourage abstention 
typically rely on calibration techniques, including linguistic calibration~\citep{mielke2022reducing, huang2024calibrating}, which aim to accurately and consistently estimate a model's confidence in its response, then arrange for the model to abstain if the confidence score for a given response falls below some threshold~\citep{varshney2022investigating, xiao2022uncertainty, desai2020calibration}. But questions of whether a query is aligned with human values or is answerable at all are difficult to model in terms of model confidence~\cite{yang2023alignment}.


Prior work demonstrates the potential of abstention to enhance model safety and reliability in constrained settings~\cite{varshney2023art, wang2024not, zhang2023r}.
In this survey, we attempt to bring together relevant work studying abstention strategies or leading to abstention behaviors, across the diverse range of scenarios encountered by general-purpose chatbots engaging in open-domain interactions.
Our goals are to identify gaps and encourage new methods to achieve abstention. Developing or adapting abstention mechanisms to suit a wide array of tasks will enhance the overall robustness and trustworthiness of LLM interactions. 

To this end, our survey presents an overview of the current landscape of abstention research. We provide a definition of abstention that incorporates not only technical perspectives---query examination and model capabilities---but also considers alignment with human values. We categorize existing methods to improve abstention in LLMs based on the model lifecycle (pretraining, alignment, and inference), and provide an analysis of evaluation benchmarks and metrics used to assess abstention.
In our discussion, we aim to establish a clear entry point for researchers to study the role of abstention across tasks, facilitating the incorporation of new abstention techniques into future LLM systems.

We summarize our contributions below:
\begin{itemize}[itemsep=-1pt, topsep=3pt, leftmargin=10pt]
\item We introduce a framework to analyze abstention capabilities from three perspectives that have typically been considered in isolation---query answerability, the confidence of the model to answer the query, and alignment of query and responses with human values. Our framework helps us identify existing research that is relevant to abstention as well as abstention mechanisms that have been developed in prior work
(\S\ref{sec:abstentain_system}).

\item We conduct a detailed survey of existing abstention methods(\S\ref{sec:method}) as well as evaluation benchmarks and metrics (\S\ref{sec:eval}), aiding researchers in selecting appropriate strategies. For each class of methods, we identify opportunities for further research to advance the field.

\item We discuss other considerations and under-explored aspects (\S\ref{sec:challenges}) of abstention, highlighting pitfalls and promising future directions. We encourage researchers to develop more robust model abstention mechanisms and demonstrate their effectiveness in real-world applications.

\end{itemize}

\vspace{-2mm}
\section{Abstention in LLMs}  
\label{sec:abstentain_system}

\vspace{-2mm}
\paragraph{Definition} \label{abs_def} We define \textit{\textbf{abstention}} as the refusal to answer a query.
When a model fully abstains, it 
may begin a response with ``I don't know'' or refuse to answer in another way. 
In reality, abstention encompasses a spectrum of behaviors
~\cite{rottger2023xstest}, e.g., expressing uncertainty, providing conflicting conclusions, or refusing due to potential harm are all forms of abstention. \textit{\textbf{Partial abstention}} may involve both answering and abstention, such as self-contradictory responses, e.g., ``I can't answer the question, but I suppose the answer might be...'' We do not consider ignoring and/or reframing the question as abstention; but rather as failure modes of LLMs in following instructions~\citep{rottger2023xstest, varshney2023art}.

For the \textit{\textbf{abstention expression}}---the words a model uses to convey that it has abstained---we adopt the definition of five major types of expressions from prior work~\citep{varshney2023art, wang2024not}, indicating that  
the model (i) cannot assist; (ii) refutes the query; (iii) provides multiple perspectives without expressing preference; (iv) perceives risk associated with the query and answers cautiously with a disclaimer; and (v) refuses to offer concrete answers due to the lack of knowledge or certainty. Expressions can be identified through heuristic rules and key word matching \citep{zou2023universal, wen-etal-2024-characterizing, yang2023alignment}, or through model-based or human-based evaluation (\S\ref{sec:eval}). 

~ \\ [-3mm]
\noindent Below, we describe and motivate our framework for analyzing abstention behavior (\S\ref{sec:abstention_framework}), then provide a formal definition of its components (\S\ref{sec:formal_definition}).

\vspace{-1mm}
\subsection{Abstention Framework}
\label{sec:abstention_framework}
\vspace{-1mm}

We study abstention in the scenario of LLMs as AI assistants, exemplified by chatbots such as
ChatGPT~\citep{chatgpt, achiam2023gpt}, Claude~\cite{claude}, LLaMA~\citep{touvron2023llama}, and others~\cite{vicuna2023}.
We propose an idealized abstention-aware workflow for these systems in Fig.~\ref{fig:system}. Given an LLM $f$ that supports arbitrary generative modeling tasks and the users' input $\mathbf{x}$, $f$ generates an output $\mathbf{y}$. We analyze the decision to abstain from three distinct but interconnected perspectives:

\begin{itemize}[itemsep=-2pt, topsep=1pt, leftmargin=10pt]
\item The \textbf{query} perspective focuses on the nature of the input---whether the query is ambiguous or incomplete~\cite{asai-choi-2021-challenges}, beyond what any human or model could possibly know~\cite{amayuelas2023knowledge}, there is irrelevant or insufficient context to answer \citep{aliannejadi2019asking, li2024mediq}, or there are knowledge conflicts \citep{Wang2023ResolvingKC}. In these situations, the system should abstain.

\item The \textbf{model knowledge} perspective examines the capabilities of the AI model itself, including its design, training, and inherent biases~\citep{ahdritz2024distinguishing, kim2024epistemology,hestness2017deep, hoffmann2022training,kaplan2020scaling, cao-2024-learn}. For any given query, the system should abstain if the model is insufficiently confident about the correctness of output
or has a high probability of returning an incorrect output.

\item The \textbf{human values} perspective considers ethical implications and societal norms that influence whether a query should be answered,
emphasizing the impact of responses on human users~\citep{kirk2023past}. 
A system should abstain if asked for personal opinions or values (i.e., the query anthropomorphizes the model), or if the query or response may compromise safety, privacy, fairness, or other values.  
\end{itemize}

\noindent For examples of queries and outputs meeting conditions for abstention, please see Appendix Tab.~\ref{tab:perspectives_examples}.

\subsection{Problem Formulation}
\label{sec:formal_definition}

To formalize our definition of abstention: consider an LLM $f: \mathcal{X} \rightarrow \mathcal{Y}$. When given a prompt $ \mathbf{x} \in \mathcal{X}$, $f$ generates a response $\mathbf{y} \in \mathcal{Y}$.  We model refusal to answer (abstention) as a function $r: \mathcal{X,Y} \rightarrow [0, 1]$ where $r(\mathbf{x},\mathbf{y})=1$ indicates the system will fully abstain from answering, $r(\mathbf{x},\mathbf{y})=0$ indicates the system will return the output $\mathbf{y}$, and intermediate values represent partial abstention.

We define $r$ as the conjunction of three functions, to be defined by a system designer, to assess query \textit{answerability},
the \textit{confidence} of the LLM's response to the query, and the \textit{human value alignment} of the query and response. 
We define these three functions as:

\begin{itemize}[itemsep=-1pt, topsep=1pt, leftmargin=10pt]
    \item Query function $a: \mathcal{X} \rightarrow [0, 1]$. $a(\mathbf{x})$ represents the degree to which an input
    $\mathbf{x}$ can be answered. 

    \item Model confidence function $c: \mathcal{X,Y} \rightarrow [0, 1]$. $c(\mathbf{x},\mathbf{y})$ indicates the model $f$'s confidence in its output $\mathbf{y}$ based on input $\mathbf{x}$.
    \item Human value alignment functions $h: \mathcal{X,Y} \rightarrow [0, 1]$. We define two variants of $h$: $h(\mathbf{x})$ operates on the input alone and determines its alignment with human values, and $h(\mathbf{x},\mathbf{y})$ operates on both the input $\mathbf{x}$ and predicted output $\mathbf{y}$. $h$ is measured either through human annotation~\citep{ouyang2022training} or a proxy model that can be learned based on human preferences~\citep{gao2023scaling}.
\end{itemize}

\noindent The refusal function $r$ determines whether the LLM should abstain from responding to input $\mathbf{x}$ as:

\setlength{\abovedisplayskip}{2pt}
\setlength{\belowdisplayskip}{4pt}
\small
\vspace{-2mm}
\begin{equation*}
    r(\mathbf{x},\mathbf{y}) =
    \begin{cases} 
    1, \textrm{if any of  } a(\mathbf{x}); c(\mathbf{x},\mathbf{y}); h(\mathbf{x},\mathbf{y}) = 0\\
    M(a(\mathbf{x}), c(\mathbf{x},\mathbf{y}), h(\mathbf{x},\mathbf{y})), \textrm{otherwise} \\
    0, \textrm{if all of  } a(\mathbf{x}); c(\mathbf{x},\mathbf{y}); h(\mathbf{x},\mathbf{y}) = 1
    \end{cases}
\end{equation*}

\normalsize

\noindent The function $M$ acts as a connector between the three perspectives, defining how the system integrates their individual outputs into a unified decision. The design of $M$ depends on the system designer and 
will vary based on the application; examples include weighted averaging, logical operations, or custom thresholds. Some existing systems~\citep{varshney2022investigating, cole2023selectively} use threshold-based mechanisms to convert partial abstention behaviors into binary decisions, such as full compliance or full refusal. 

Our framework allows nuanced handling of abstention, by combining confidence from all three perspectives and
enabling partial abstention when appropriate. 
Under this definition, 
a system would fully abstain from answering if any of the three perspectives indicates full abstention. In all other cases, a system would partially abstain, balancing between providing an answer and withholding information based on 
indications from the three perspectives. 


\subsection{Inclusion in this Survey}

We identify and survey prior work that falls under any of the three perspectives of our abstention framework. In \S\ref{sec:method}, we organize abstention methodology from an \textit{LLM-centered} perspective, based on when each method is applied in the LLM lifecycle:
pretraining, alignment, or inference. This organization is chosen for ease of comparison of experimental settings. Each subsection within \S\ref{sec:method} is further organized by the three perspectives. Following, \S\ref{sec:eval}  describes evaluation benchmarks and metrics that have been used or introduced across the surveyed prior work. At the end of each subsection, in blue boxes, we summarize main takeaways and provide suggested directions for future work. In \S\ref{sec:challenges}, we summarize notable threads of research that are not easily classified as method or evaluation.

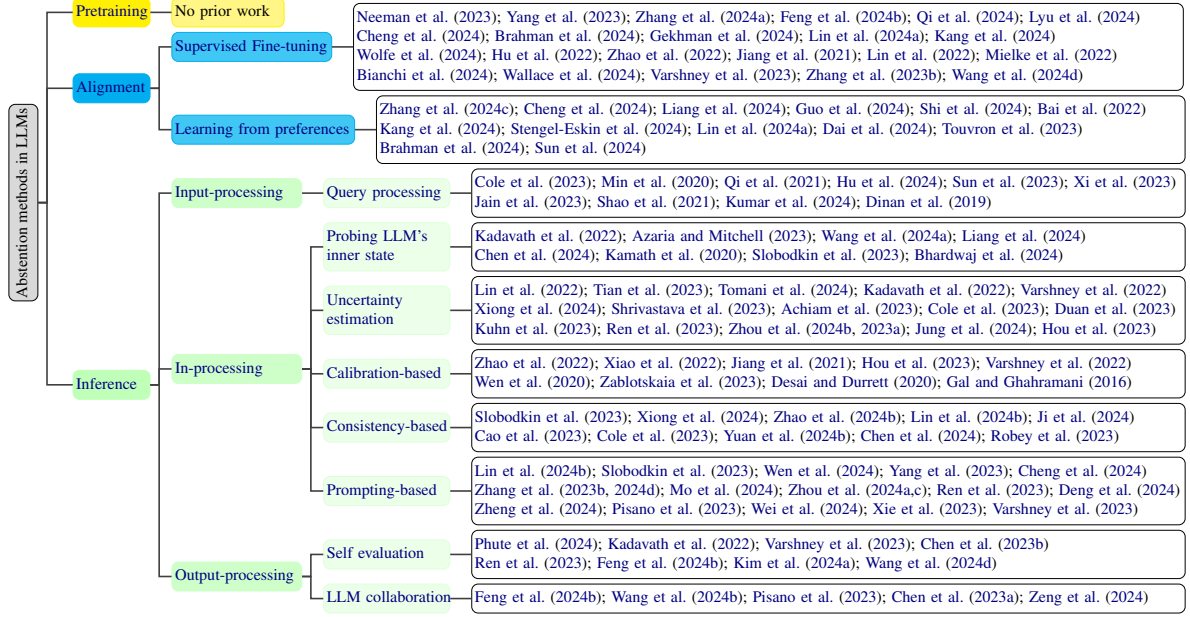
\begin{figure*}[ht!]
    \centering
    \resizebox{0.98\textwidth}{!}{
        \begin{forest}
            forked edges,
            for tree={
                grow=east,
                reversed=true,
                anchor=base west,
                parent anchor=east,
                child anchor=west,
                base=left,
                font=\small,
                rectangle,
                draw=black,
                rounded corners,
                align=left,
                minimum width=4em,
                edge+={darkgray, line width=1pt},
                s sep=3pt,
                inner xsep=2pt,
                inner ysep=3pt,
                ver/.style={rotate=90, child anchor=north, parent anchor=south, anchor=center},
            },
            [
                Abstention methods in LLMs, 
                ver, color=black!100, fill=black!15, text=black
                [
                    {\hyperref[methods:pretraining]{Pretraining}}, 
                    color=yellow!100, fill=yellow!100, text width=4.0em, text=black
                    [
                        {No prior work}, 
                        color=yellow!100, fill=yellow!60,  text width=6em, text=black
                    ]
                ]
                [
                    {\hyperref[methods:alignment]{Alignment}}, color=cyan!100, fill=cyan!100, text width=4.0em, text=black
                    [
                        {\hyperref[methods:instruction_tuning]{Supervised Fine-tuning}}, 
                        color=cyan!100, fill=cyan!60,  text width=8.7em, text=black
                        [
                            \citet{neeman2022disentqa, yang2023alignment, zhang2023r,feng2024don, qi2023fine, lyu2024keeping} \\ 
                         \citet{cheng2024ai, brahman-kumar2024, gekhman2024does, lin2024flame, kang2024unfamiliar} \\
                              \citet{10.1145/3630106.3658966, hu2021lora, zhao2022calibrating, jiang2021can, lin2022teaching, mielke2022reducing} \\ 
                            \citet{bianchi2023safety,wallace2024instruction, varshney2023art, zhang2023defending, wang2024selfguard},
                            text width=46.7em
                        ]                        
                    ]
                    [
                        {\hyperref[methods:learning_preferences]{Learning from preferences}}, 
                        color=cyan!100, fill=cyan!60,  text width=10em, text=black
                        [
                            \citet{zhang2024selfalignment, cheng2024ai, liang2024learning, guo2024controllable,shi2024saferinstruct,bai2022training} \\
                            \citet{kang2024unfamiliar, stengel2024lacie, lin2024flame, dai2023safe, touvron2023llama} \\
                            \citet{brahman-kumar2024,sun2024salmon},                            
                            text width=45.5em
                        ]
                    ]
                ]
                [
                    {\hyperref[methods:inference]{Inference}}, color=lightgreen!200, fill=lightgreen!200, text width=4.0em, text=black
                    [
                        {\hyperref[methods:input_processing]{Input-processing}}, 
                        color=lightgreen!150, fill=lightgreen!150, text width=7em, text=black  
                        [
                            {\hyperref[methods:query_processing]{Query processing}},
                            color=lightgreen!100, fill=lightgreen!60,  text width=6.8em, text=black
                            [
                                \citet{cole2023selectively, min2020ambigqa, qi2020onion, hu2024tokenlevel, sun2023defending, xi2024defending} \\
                                \citet{jain2023baseline, shao2021bddr, kumar2024certifying, dinan2019build}, 
                                text width=40.1em
                            ]
                        ]
                    ]
                    [
                        {\hyperref[methods:in_processing]{In-processing}}, 
                        color=lightgreen!150, fill=lightgreen!150, text width=7em, text=black
                        [
                            {\hyperref[methods:probing_llm_state]{Probing LLM’s}} \\ {\hyperref[methods:probing_llm_state]{inner state}}, 
                            color=lightgreen!100, fill=lightgreen!60,  text width=6.8em, text=black
                            [
                                \citet{kadavath2022language, azaria2023internal, wang2024inferaligner, liang2024learning} \\
                                \citet{chen2024inside, kamath2020selective, slobodkin-etal-2023-curious,bhardwaj2024language},
                                text width=40.1em
                            ]
                        ]
                        [
                            {\hyperref[methods:uncertainty_estimation]{Uncertainty}} \\ {\hyperref[methods:uncertainty_estimation]{estimation}}, 
                            color=lightgreen!100, fill=lightgreen!60,  text width=6.8em, text=black
                            [
                                \citet{lin2022teaching, tian2023just,tomani2024uncertainty,kadavath2022language,varshney2022investigating} \\
                                \citet{xiong2023can,shrivastava2023llamas,achiam2023gpt, cole2023selectively, duan2023shifting} \\
                                \citet{kuhn2023semantic, ren2023self, zhou2024relying, zhou2023navigating, jung2024trust, hou2023decomposing},
                                text width=40.1em
                            ]
                        ]
                        [
                            {\hyperref[methods:calibration]{Calibration-based}}, 
                            color=lightgreen!100, fill=lightgreen!60,  text width=6.8em, text=black
                            [
                                \citet{zhao2022calibrating,xiao2022uncertainty,jiang2021can, hou2023decomposing, varshney2022investigating} \\ 
                                \citet{wen2020batchensemble, zablotskaia2023uncertainty,desai2020calibration, pmlr-v48-gal16}, 
                                text width=40.1em
                            ]
                        ]
                        [
                            {\hyperref[methods:consistency]{Consistency-based}}, 
                            color=lightgreen!100, fill=lightgreen!60,  text width=6.8em, text=black
                            [
                                \citet{slobodkin-etal-2023-curious,xiong2023can, zhao2023knowing,lin2024generating,ji2024defending} \\
                                \citet{cao2023defending,cole2023selectively,yuan2024rigorllm,chen2024inside,robey2023smoothllm}, 
                                text width=40.1em
                            ]
                        ]
                        [
                            {\hyperref[methods:incontext_learning]{Prompting-based}}, 
                            color=lightgreen!100, fill=lightgreen!60,  text width=6.8em, text=black
                            [
                                \citet{lin2024generating, slobodkin-etal-2023-curious, wen-etal-2024-characterizing, yang2023alignment,cheng2024ai} \\
                          \citet{zhang2023defending,zhang2024intention, mo2024fight,zhou2024robust, zhou2024defending, ren2023self, deng-etal-2024-dont} \\
                                \citet{zheng2024prompt,pisano2023bergeron,wei2024jailbreak,xie2023defending,varshney2023art}, 
                                text width=40.1em
                            ]
                        ]
                    ]
                    [
                        {\hyperref[methods:output_processing]{Output-processing}}, 
                        color=lightgreen!150, fill=lightgreen!150, text width=7em, text=black
                        [
                            {\hyperref[methods:self_evaluation]{Self evaluation}}, 
                            color=lightgreen!100, fill=lightgreen!60,  text width=6.8em, text=black
                            [
                                \citet{helbling2023llm, kadavath2022language, varshney2023art, chen2023adaptation} \\
                                \citet{ren2023self, feng2024don, kim2024break, wang2024selfguard}, 
                                text width=40.1em
                            ]
                        ]
                        [
                            {\hyperref[methods:llm_collaboration]{LLM collaboration}}, 
                            color=lightgreen!100, fill=lightgreen!60,  text width=6.8em, text=black
                            [
                                \citet{feng2024don, wang2024defending, pisano2023bergeron, chen2023jailbreaker,zeng2024autodefense}, 
                                text width=40.1em
                            ]
                        ]
                    ]
                ]
            ]
        \end{forest}
    }
\caption{Methods to improve LLM abstention grouped by pretraining, alignment, and inference stages.}
\label{fig:categorization_of_survey}
\end{figure*}

\section{Abstention Methodology} 

\label{sec:method}

We summarize methods introduced in prior work  (Fig.~\ref{fig:categorization_of_survey} organizes these by stages in the LLM lifecycle) and provide ideas for future experiments.

\subsection{Pretraining Stage}
\label{methods:pretraining}
We found no existing research that studies abstention in the pretraining stage, despite the widely recognized importance of pretraining as a critical phase for model knowledge acquisition. To bridge this gap, we propose several directions for future exploration.

\vspace{1mm}
\noindent \colorbox{blue!8}{
\begin{minipage}{0.46\textwidth}
\footnotesize
{\color{blue}{\textbf{Pretraining Summary}}}

\begin{itemize}[itemsep=-1pt, topsep=2pt, leftmargin=6pt, labelsep=3pt]
\item The impact of refusal-aware data in pretraining is not well-studied.
\setlength{\labelsep}{1pt}
\item[\idea] Demystify Pretraining Corpora: Investigate how the distribution of refusal-aware data in pretraining corpora impact model abstention ability.
\item[\idea] Coarse-to-fine Abstention Pretraining: 
In later stages of pretraining, adding data source information including the domain and reasons to abstain into pretraining may be helpful for improving abstention ability.
\item[\idea] Curriculum Learning for Abstention: Simpler abstention scenarios can be introduced first in pretraining, progressing to more complex or fine-grained abstention tasks in later pretraining phases.
\item[\idea] Hybrid Objective Function: Modify the next-word prediction objective in later pretraining stages to incorporate an abstention-aware loss component, balancing between generating tokens and abstaining when appropriate.
\item[\idea] Regularization Techniques: Apply abstention-aware regularization during later stages of pretraining to encourage the model to abstain when prediction uncertainty is high.

\end{itemize}

\end{minipage}} \\ [-2mm]



\subsection{Alignment Stage}
\label{methods:alignment}
\vspace{-1mm}
We categorize alignment-stage methods as supervised finetuning (SFT) or preference optimization (PO). Some papers include both methods, as PO usually requires SFT as a precursor; we discuss these in the subsection most reflective of their primary contributions.

\paragraph{Supervised Finetuning\label{methods:instruction_tuning}} Many works have demonstrated that SFT with abstention-aware data can improve model abstention capabilities. For example, \citet{neeman2022disentqa} perform data augmentation in the finetuning stage to encourage LLMs to predict unanswerable when presented with an empty or randomly sampled document. \citet{yang2023alignment} construct an honesty alignment dataset by substituting LLM's wrong or uncertain responses with ``I don't know'' and finetuning on the resulting data, improving model abstention. Notably, \citet{zhang2023r} introduce $R$-tuning, constructing and finetuning on a refusal-aware dataset and showing improved abstention capabilities. \citet{zhang2023r} also argue that refusal-aware answering is task-independent and could benefit from multi-task training and joint inference. However, \citet{feng2024don} present contradictory findings in their Appendix that abstention-aware instruction-tuning struggles to generalize across domains and LLMs. 

In parallel, concerns have emerged regarding the effectiveness of SFT for abstention. \citet{cheng2024ai} and \citet{brahman-kumar2024} find that SFT can make models more conservative, leading to a higher number of incorrect refusals. Recent work~\citep{gekhman2024does, lin2024flame, kang2024unfamiliar} also demonstrates that finetuning on examples unobserved during pretraining increases the risk of hallucination. \citet{gekhman2024does} propose a mitigation strategy to re-label these examples based on the pretrained LLM’s knowledge and include ``I don’t know'' in the finetuning data to teach the model to abstain. A related method to reduce hallucination is introduced in~\citet{lin2024flame}; the authors create factuality-aware training data for SFT by classifying whether an instruction requires a factual response. 

Parameter-efficient finetuning (PEFT) strategies have also been used for abstention. \citet{10.1145/3630106.3658966} conduct lab-scale experiments, finetuning LLMs with QLoRA~\citep{NEURIPS2023_1feb8787}, and observe that weaker models (with lower task performance) tend to achieve greater gains in abstention performance. Beyond resource efficiency, \citet{brahman-kumar2024} have found that LoRA~\citep{hu2021lora} acts as an effective regularization technique for improving abstention; they find that fully finetuned models exhibit over-refusal while also forgetting general capabilities, and demonstrate that finetuning with LoRA alleviates both issues while significantly improving abstention behavior.

Instead of finetuning directly, finetuning for calibration may indirectly improve abstention ability \citep{Szegedy_2016_CVPR, zhao2022calibrating, xiao2022uncertainty, jiang2021can, lin2022teaching}. \citet{jiang2021can} propose two finetuning objective functions (softmax-based and margin-based), which improve ECE~\citep{guo2017calibration} on multiple-choice datasets. \citet{mielke2022reducing} alternatively use a calibrator trained to provide a confidence score with an LLM finetuned to control for linguistic confidence in a system.

Towards alignment with human values, \citet{bianchi2023safety} show that adding a small number of safety instructions to instruction-tuning data reduces harmful responses without diminishing general capabilities, whereas an excessive number of safety instructions makes LLMs overly defensive. \citet{varshney2023art} construct responses for unsafe prompts by combining fixed refusal responses with Llama-2-generated safe responses, and obtain similar results. \citet{wallace2024instruction} finetune LLMs to follow hierarchical prompts,
enhancing the fine-grained abstention ability of LLMs. \citet{zhang2023defending} also finetune LLMs with goal prioritization instructions that instruct LLMs to prioritize safety over helpfulness during inference. 

However, custom finetuning of LLMs presents safety risks. For example, \citet{qi2023fine,lyu2024keeping} note that finetuning with benign and commonly used datasets increases unsafe behaviors in aligned LLMs \citep{qi2023fine}. To address this, \citet{lyu2024keeping} propose to finetune models without a safety prompt, but include one at test time. \citet{wang2024selfguard} finetune LLMs to evaluate their own outputs for harm and append a ``harmful'' or ``harmless'' tag to its responses instead of directly tuning LLMs to abstain.

\vspace{1mm}
\noindent \colorbox{blue!8}{
\begin{minipage}{0.46\textwidth}
\footnotesize
{\color{blue}{\textbf{Instruction Tuning Summary}}}

\begin{itemize}[itemsep=-1pt, topsep=2pt, leftmargin=6pt, labelsep=3pt]
\item Instruction tuning on abstention-aware data improves abstention ability but can lead to over-abstention.

\item Researchers disagree on whether instruction tuning helps LLMs learn abstention as a meta-capability.

\setlength{\labelsep}{1pt}
\item[\idea] Heterogeneous approaches: Abstention-aware instruction-tuning datasets tend to emphasize only a few question-response formats. Incorporating diverse abstention expressions, definitions, and domains may improve generalization. Ensembling models trained in different abstention domains, e.g., mixture of LoRA experts, may also help achieve better generalization.
\end{itemize}
\end{minipage}} \\  [-2mm]

\paragraph{Learning from preferences\label{methods:learning_preferences}} 
Preference optimization can impact abstention from both the model knowledge and human value alignment perspectives.
As described above, finetuning LLMs on abstention-aware data may lead to overly conservative behavior, causing erroneous refusals of queries. \citet{cheng2024ai} and \citet{brahman-kumar2024} address this through Direct Preference Optimization (DPO)~\citep{rafailov2023direct}, encouraging the model to answer questions it knows and refuse questions it does not know.  

Factuality-based preference optimization can help models respond correctly to queries, including abstaining (e.g., saying ``I don't know''). 
As an example, \citet{liang2024learning} construct a factual preference dataset to train a reward model, and utilize it to optimize abstention preferences in LLMs via Proximal Policy Optimisation (PPO)~\citep{schulman2017proximal}. \citet{kang2024unfamiliar} design a reward function that prioritizes abstention over incorrect answers, while \citet{lin2024flame} incorporate factuality-focused preference pairs into DPO to enhance fact-based instruction following. 

Other works use DPO to improve calibration, which can also aid abstention. LACIE~\citep{stengel2024lacie} casts confidence calibration as a preference optimization problem and introduce a speaker-listener game to create preference data; they demonstrate that finetuning on LACIE data leads to emergent model abstention behavior. \citet{zhang2024selfalignment} introduce Self-Alignment for Factuality, generating confidence scores through self-asking to improve calibration via DPO.

For human values, safety alignment methods \citep{dai2023safe, touvron2023llama, bai2022training, shi2024saferinstruct} use explicit or implicit preference models to reduce harmfulness, which though not explicitly focused on abstention, will encourage abstention on unsafe prompts. Other studies have explored multi-objective alignment approaches~\citep{guo2024controllable} to encourage safe and helpful model behavior. The instructable reward model in SALMON \citep{sun2024salmon} is trained on synthetic preference data, generating reward scores based on customized human-defined principles as the preference guideline.

\vspace{1mm}
\noindent \colorbox{blue!8}{
\begin{minipage}{0.46\textwidth}
\footnotesize
 {\color{blue}{\textbf{Learning from Preferences Summary}}}
 
 \begin{itemize}[itemsep=-1pt, topsep=2pt, leftmargin=6pt, labelsep=3pt]

\item Preference optimization can reduce over-abstention caused by instruction tuning on refusal-aware data by aligning model behavior with user preferences. However, this may not always be the case. Preference optimization may still lead to over-abstention if reward models overemphasize safety or preference data favors abstention. Balancing feedback and ensuring diverse preference data are crucial to address this challenge, making it an important area for further research.



\setlength\labelsep{1pt}
\item[\idea] Ranking-based preference optimization: Since abstention is a spectrum of behaviors, ranking-based preference optimization can extend the pair-wise contrast to accommodate rankings over different degrees of abstention. 


\end{itemize}

\end{minipage}} \\  [-2mm]

\subsection{Inference Stage\label{methods:inference}}
 
We categorize inference stage methods as input-processing, in-processing, or output-processing approaches based on when they are applied. Input-processing approaches are centered on the query answerability and human values perspectives; in-processing approaches on the model knowledge perspective; and output-processing approaches may consider both model knowledge and human values.  

\subsubsection{Input-processing Approaches}
\label{methods:input_processing}

\paragraph{Query processing\label{methods:query_processing}}
From the query perspective in our proposed framework, LLMs can choose to abstain based on the query answerability. For example, \citet{cole2023selectively} try to predict the ambiguity of questions derived from the AmbigQA dataset \citep{min2020ambigqa} before selectively answering.

Other methods aim to identify queries that are misaligned with human values. For example, \citet{qi2020onion} detect malicious queries needing abstention by removing suspect words from the query and analyzing the resulting drop in perplexity while \citet{hu2024tokenlevel} propose new ways of computing perplexity and find tokens with abnormally high perplexity. Apart from perplexity-based methods, \citet{jain2023baseline} further investigate input preprocessing methods such as paraphrasing and retokenization. The BDDR framework~\citep{shao2021bddr} not only detects suspicious words in the input but also reconstructs the original text through token deletion or replacement.
\citet{kumar2024certifying} introduce the ``erase-and-check'' framework to defend against adversarial prompts with certifiable safety guarantees. 
Similarly, \citet{xi2024defending} measure changes in representation between original and paraphrased queries using a set of distributional anchors to identify harmful queries. \citet{dinan2019build} develop a more robust offensive language detection system through an iterative build-it, break-it, fix-it strategy.

\vspace{1mm}
\noindent \colorbox{blue!8}{
\begin{minipage}{0.46\textwidth}
\footnotesize
 {\color{blue}{\textbf{Query Processing Summary}}}
 
 \begin{itemize}[itemsep=-1pt, topsep=2pt, leftmargin=6pt, labelsep=3pt]
\item  Query processing approaches focus on assessing query ambiguity or human values alignment.

\setlength\labelsep{1pt}
\item[\idea] Context awareness: Existing work focuses on query-only processing or simple context-dependent tasks where insufficient or conflicting context may be provided in the abstention scenario~\citep{rajpurkar2018know, kwiatkowski2019natural}. These settings overlook context complexity in real-world applications; for example, earlier context in multi-turn conversations can impact judgments for either query answerability or human values alignment in later conversational turns.

\end{itemize}
 
\end{minipage}} \\ [-2mm]

\subsubsection{In-processing Approaches}
\label{methods:in_processing}

\paragraph{Probing LLM's inner state\label{methods:probing_llm_state}} 

Recent studies \citep{kamath2020selective,azaria2023internal} focus on training calibrators based on LLM internal representation to predict the accuracy of the model’s responses, enabling abstention when the likelihood of error is high. Further probing into the internal representations of LLMs to discern between answerable and unanswerable queries has been conducted by \citet{slobodkin-etal-2023-curious}, \citet{kadavath2022language} and \citet{liang2024learning}. Additionally, \citet{chen2024inside} introduce the EigenScore, a novel metric derived from LLM's internal states, which can facilitate abstention by quantifying the reliability of the model's knowledge state.

In terms of leveraging the LLMs' internal states for safety judgments, \citet{wang2024inferaligner} extract safety-related vectors (SRVs) from safety-aligned LLMs; which are then used as an abstention gate to steer unaligned LLMs towards safer task performance. Furthermore, \citet{bhardwaj2024language} demonstrate that integrating a safety vector into the weights of a finetuned LLM through a simple arithmetic operation can significantly mitigate the potential harmfulness of the model’s responses.

\vspace{1mm}
\noindent \colorbox{blue!8}{
\begin{minipage}{0.46\textwidth}
\footnotesize
 {\color{blue}{\textbf{Probing LLM's Inner State Summary}}}
 
 \begin{itemize}[itemsep=-1pt, topsep=2pt, leftmargin=6pt, labelsep=3pt]
\item  LLM's internal representations can indicate model knowledge boundaries and safety awareness.
\end{itemize}
 
\end{minipage}}

\paragraph{Uncertainty estimation\label{methods:uncertainty_estimation}} Estimating the uncertainty of LLM output can serve as a proxy for making abstention decisions. Token-likelihoods have been widely used to assess the uncertainty of LLM responses~\citep{lin2022teaching, kadavath2022language}. Enhancing this approach, \citet{lin2022teaching} and \citet{tian2023just} employ an indirect logit methodology to calculate the log probability of the ‘True’ token when appended to model’s generated response. \citet{shrivastava2023llamas} leverage a surrogate LLM with access to internal probabilities to approximate the confidence of the original model. \citet{tomani2024uncertainty} assess Predictive Entropy and Semantic Entropy~\citep{kuhn2023semantic} of responses. \citet{duan2023shifting} design a weighted Predictive Entropy by considering the relevance of each token in reflecting the semantics of the whole sentence. However, other work shows that aligned LLMs may not have well-calibrated logits~\citep{cole2023selectively, achiam2023gpt} and may have positional bias and probability dispersion~\citep{ren2023self}. In the context of LLM-as-judge, these canonical probability-based methods tend to be overconfident in estimating agreement with the majority of annotators; \citet{jung2024trust} propose a novel confidence estimation method by simulating diverse annotator preferences with in-context learning.


The Maximum Softmax Probability approach \citep{varshney2022investigating} uses peak softmax output as a uncertainty estimator. \citet{hou2023decomposing} introduce an uncertainty estimation method: input clarification ensembling;
through ruling out aleatoric uncertainty by clarification, the remaining uncertainty of each individual prediction is epistemic uncertainty. 

Beyond probability-based measures, verbalized confidence scores have emerged as another class of methods to estimate and manage uncertainty~\citep{lin2022teaching, tian2023just, tomani2024uncertainty, xiong2023can,zhou2024relying}. \citet{xiong2023can} examine prompting methods including chain-of-thought~\citep{NEURIPS2022_9d560961}, self-probing, top-$k$~\citep{tian2023just}, and linguistic likelihood expressions to eliciting confidence scores. Although LMs can be explicitly prompted to
express confidence, verbalized confidence scores have been found to be over-confident~\citep{xiong2023can, zhou2024relying}. \citet{zhou2024relying} find that LMs are reluctant to express uncertainty when answering questions, even when their responses are incorrect. \citet{zhou2023navigating} show that high-certainty expressions in the prefix of a response can result in accuracy drop compared to low-certainty expressions, suggesting that LLMs respond more to prompting style rather than accurately assessing epistemic uncertainty.

\vspace{1mm}
\noindent \colorbox{blue!8}{
\begin{minipage}{0.46\textwidth}
\footnotesize
{\color{blue}{\textbf{Uncertainty Estimation Summary}}}
\begin{itemize}[itemsep=-1pt, topsep=2pt, leftmargin=6pt, labelsep=3pt]
\item Uncertainty can act as proxy for abstention. However, neither probability-based measures nor verbalized confidence may be well-calibrated.

\setlength{\labelsep}{1pt}
\item[\idea]Uncertainty estimation could be used as a key aspect of explainable AI. The model could communicate its uncertainty or insufficient confidence in its response, which could foster user trust and engagement by making the model's decision-making process more transparent.  
\end{itemize}
 
\end{minipage}}

\paragraph{Calibration-based methods\label{methods:calibration}} 

Estimated model uncertainty may not accurately represent the likelihood of a model's outputs being correct,
so numerous studies focus on calibrating the uncertainty of LLMs.
\citet{jiang2021can} improve calibration by augmenting inputs and paraphrasing outputs. Temperature Scaling \citep{pmlr-v70-guo17a, xiao2022uncertainty, desai2020calibration, jiang2021can} modifies the softmax temperature to refine calibration during decoding. Additionally, Monte-Carlo Dropout \citep{pmlr-v48-gal16, varshney2022investigating, zablotskaia2023uncertainty} employs multiple predictions with varying dropout configurations to assemble a robust confidence estimate. Batch Ensemble~\citep{wen2020batchensemble} is a computationally efficient method that aggregates multiple model predictions and maintains good calibration.



\vspace{1mm}
\noindent \colorbox{blue!8}{
\begin{minipage}{0.46\textwidth}
\footnotesize
 {\color{blue}{\textbf{Calibration-based Methods Summary}}}
 
\begin{itemize}[itemsep=-1pt, topsep=2pt, leftmargin=6pt, labelsep=3pt]

\item Confidence calibration is a longstanding area of research; methods developed for calibration can also enhance abstention capabilities, improving model output reliability.

\setlength\labelsep{1pt}
\item[\idea] Domain disparity: Investigating methods to ensure calibration techniques remain effective across diverse datasets and different types of models, particularly under domain shifts, can surface task-specific optimizations.
\end{itemize}
\end{minipage}}

\paragraph{Consistency-based methods\label{methods:consistency}} Given the limitations of confidence elicitation, some methods leverage consistency-based aggregation to estimate LLM uncertainty and then abstain when uncertain. Aggregation can be achieved using diversity and repetition~\citep{cole2023selectively}, weighted confidence scores and pairwise ranking~\citep{xiong2023can}, or semantic similarity between responses~\citep{lin2024generating, zhao2023knowing,chen2024inside}. \citet{slobodkin-etal-2023-curious} relax beam search and abstain if any top-$k$ answer is ``unanswerable''. 

Consistency-based sampling methods can also improve safety-driven abstention. \citet{robey2023smoothllm}, \citet{cao2023defending}, and \citet{ji2024defending} perturb inputs with character masks, insertions, deletions, or substitutions, and identify inconsistencies among responses, which suggest the presence of an attack prompt needing abstention. \citet{yuan2024rigorllm} obtain samples by prompting for augmentations (learnable safe suffixes and paraphrasing) and use a kNN-based method to aggregate responses.

\vspace{1mm}
\noindent \colorbox{blue!8}{
\begin{minipage}{0.46\textwidth}
\footnotesize
 {\color{blue}{\textbf{Consistency-based Methods Summary}}}
 
\begin{itemize}[itemsep=-1pt, topsep=2pt, leftmargin=6pt, labelsep=3pt]

\item Consistency-based methods establish model certainty based on its output distribution, helping to identify queries for which a model should abstain.

\end{itemize}
\end{minipage}}

\vspace{-4mm}
\paragraph{Prompting-based methods\label{methods:incontext_learning}}

In-context examples and hints can enhance model performance on abstention. Some use few-shot exemplars of abstained and answered responses \citep{slobodkin-etal-2023-curious, varshney2023art,wei2024jailbreak}, while others incorporate instruction hints  (e.g., ``Answer the question only if answerable'' or ``Answer the below question if it is safe to answer'') \citep{wen-etal-2024-characterizing, yang2023alignment, cheng2024ai, slobodkin-etal-2023-curious}. For multiple-choice QA, adding ``None of the above'' as an answer option has been shown to be effective \citep{ren2023self, lin2024generating}. \citet{zhang2023defending} explicitly prompt LLMs to prioritize safety over helpfulness. \citet{deng-etal-2024-dont} also propose that providing explanations on the unanswerability of questions not only improves model explainability, but can produce more accurate responses.


Other work focuses on carefully designed prompts. \citet{mo2024fight} concatenate a protective prefix from attack-defense interactive training with the user query. Similarly, \citet{zhou2024robust} append trigger tokens to ensure safe outputs under adversarial attacks. \citet{pisano2023bergeron} use another LLM to add conscience suggestions to the prompt. \citet{zhang2024intention} prompt LLMs to analyze input intent and abstain if malicious. \citet{xie2023defending} incorporate self-reminders in prompts to defend against attacks, while \citet{zhou2024defending} propose Robust Prompt Optimization to improve abstention performance against adaptive attacks. \citet{zheng2024prompt} find that safety prompts can safeguard LLMs against harmful queries and further propose a safety prompt optimization method to shift query representations toward or away from the refusal direction based on query harmfulness.

\vspace{1mm}
\noindent \colorbox{blue!8}{
\begin{minipage}{0.46\textwidth}
\footnotesize
 {\color{blue}{\textbf{Prompting-based Methods Summary}}}
 
 \begin{itemize}[itemsep=-1pt, topsep=2pt, leftmargin=6pt, labelsep=3pt]
\item  Prompting with abstention examples demonstrates potential in enhancing the abstention capabilities of LLMs.

\setlength\labelsep{1pt}
\item[\idea] Interpretable prompting methods: Developing methods to effectively use instructions or choose demonstrations that capture various abstention behaviors remains under-explored, along with identifying the limits of what can be achieved through in-context learning.
\end{itemize}
\end{minipage}}

\subsubsection{Output-processing Approaches}
\label{methods:output_processing}

\paragraph{Self evaluation\label{methods:self_evaluation}} \citet{chen2023adaptation} use Soft Prompt Tuning to learn self-evaluation parameters for various tasks. However, directly asking LLMs to evaluate if their responses are certain or safe (usually in a different conversation), and to abstain if they are not, has proven effective in improving LLM abstention~\citep{helbling2023llm, kadavath2022language, varshney2023art, ren2023self, feng2024don}. \citet{kim2024break} allow the LLM to iteratively provide feedback on its own responses and refine its answers; this method achieves improvements in safety even in non-safety-aligned LLMs.
\citet{wang2024selfguard} enable LLMs to self-evaluate responses and append a [harmful] or [harmless] tag to each response; however, this approach may encourage over-abstention.

\paragraph{LLM collaboration\label{methods:llm_collaboration}}
Multi-LLM systems are effective in producing better overall responses, including improved abstention behavior. In 2-LLM systems, a test LLM is employed to examine the output of the first LLM and helps with abstaining.  In \citet{wang2024defending}, the test LLM is used to guess the most likely harmful query from the output and abstains if a harmful query is detected. \citet{pisano2023bergeron} critique and correct a model's original compliant response using a secondary LLM.

Multi-LLM systems beyond two LLMs leverage different LLMs as experts to compete or cooperate to reach a final abstention decision \citep{feng2024don, chen2023jailbreaker}. As an example, \citet{zeng2024autodefense} employ a group of LLMs in a system with an intention analyzer, original prompt analyzer, and judge.

\vspace{1mm}
\noindent \colorbox{blue!8}{
\begin{minipage}{0.46\textwidth}
\footnotesize
 {\color{blue}{\textbf{Self Evaluation \& LLM Collaboration Summary}}}
 
\begin{itemize}[itemsep=-1pt, topsep=2pt, leftmargin=6pt, labelsep=3pt]
\item LLMs can struggle with self-evaluation, but combining multiple LLMs has been effective in enhancing abstention capabilities.

\end{itemize}
\end{minipage}}

\section{Evaluation of Abstention}
\label{sec:eval}
We survey evaluation benchmarks (\S\ref{sec:evaluation_benchmarks}) and metrics (\S\ref{sec:evaluation_metrics}) used to assess abstention capabilities.

\subsection{Evaluation Benchmarks}
\label{sec:evaluation_benchmarks}

Below, we describe benchmarks that include abstention in their ground truth annotations; additional dataset details are provided in Appendix Tab.~\ref{tab:all_benchmarks}. Most evaluation datasets focus on assessing specific aspects of abstention according to our framework, though recent work from \citet{brahman-kumar2024} espouse a holistic evaluation strategy. 

\paragraph{Query-centric abstention datasets} Prior work introduces datasets containing unanswerable questions.
\squad~\citep{rajpurkar2018know} first includes unanswerable questions with \emph{irrelevant} context passages for machine reading comprehension. Rather than modifying questions to be unanswerable as in \squad, unanswerable questions in Natural Questions~\citep{kwiatkowski2019natural} are paired with insufficient context. MuSiQue~\cite{trivedi2021musique} is a multi-hop QA benchmark containing unanswerable questions for which supporting paragraphs have been removed.  CoQA~\cite{reddy-etal-2019-coqa} and QuAC~\cite{choi-etal-2018-quac} introduce unanswerable questions for conversational QA. 
Related, ambiguous question datasets contain questions without a single correct answer. AmbigQA~\citep{min2020ambigqa} extracts questions from NQ-Open~\citep{kwiatkowski2019natural} with multiple possible answers. SituatedQA~\citep{zhang-choi-2021-situatedqa} is an open-domain QA dataset where answers to the same question may change depending on when and where the question is asked. SelfAware~\citep{yin2023large} and Known Unknown Questions~\citep{amayuelas2023knowledge} consist of unanswerable questions from diverse categories. 


Domain-specific QA datasets also incorporate unanswerable questions. \pubmedqa~\citep{jin2019pubmedqa} contains biomedical questions that can be answered ``yes'', ``no'', or ``maybe''; where ``maybe'' indicates \emph{high uncertainty} based on the given context. In \qasper~\citep{dasigi2021dataset}, unanswerable questions are expert-labeled and mean that \emph{no answer is available} in the given context. 

\paragraph{Model knowledge-centric abstention datasets} RealTimeQA~\cite{kasai2022realtime} is a dynamic dataset which announces questions and evaluates systems on a regular basis, and contains inquiries about current events. PUQA (Prior Unknown QA)~\cite{yang2023alignment} comprises questions about scientific literature from 2023, beyond the cutoff of the tested models' existing knowledge.  ElectionQA23~\cite{feng2024don} is a QA dataset focusing on 2023 elections around the globe; due to the temporality of training data, LLMs lack up-to-date information to accurately respond to these queries. Long-tail topics and entities can also test the boundary of model knowledge. For example, datasets like POPQA~\citep{mallen2022not} or EntityQuestions~\citep{sciavolino-etal-2021-simple} cover knowledge on long-tail entities, which are useful for probing model knowledge boundaries. 

\vspace{-5mm}
\paragraph{Human value-centric abstention datasets} 
Here are datasets designed to measure whether LLM outputs are ``safe,'' i.e., align with widely held ethical values; these datasets may consist of prompts that are either inherently unsafe or likely to elicit unsafe responses from LLMs. Some datasets focus on specific aspects of safety. A main concern is toxicity, when models generate harmful, offensive, or inappropriate content. For instance, RealToxicityPrompts~\citep{gehman2020realtoxicityprompts} gathers prompts to study toxic language generation, while ToxiGen~\citep{hartvigsen2022toxigen} and LatentHatred~\citep{elsherief2021latent} address implicit toxic speech, and ToxicChat~\citep{lin2023toxicchat} collects data from real-world user-AI interactions. Beyond toxicity, Beavertails~\citep{ji2024beavertails} balances safety and helpfulness in QA, CValues~\citep{xu2023cvalues} assesses safety and responsibility, and Xstest~\citep{rottger2023xstest} 
examines exaggerated safety behaviors. 
LatentJailbreak~\citep{qiu2023latent} introduces a benchmark that assesses both the safety and robustness of LLMs. Do-Anything-Now~\citep{shen2023anything} collects a set of unsafe prompts for malicious purposes. 

Comprehensive safety benchmarks attempt to encompass a range of concerns. \citet{rottger2024safetyprompts} conduct the first systematic review of open datasets for evaluating LLM safety. Do-Not-Answer~\citep{wang2024not} 
includes instructions covering information hazards, malicious uses, discrimination, exclusion and toxicity, misinformation harms, and human-computer interaction harms. XSafety~\cite{wang2023all} provides a multilingual benchmark covering 14 safety issues across 10 languages. SALAD-Bench~\citep{li2024salad} is a large-scale dataset with a three-tier taxonomy, evaluating LLM safety and attack-defense methods. SORRY-Bench~\citep{xie2024sorry} proposes a more fine-grained taxonomy and diverse instructions. Most relevant to abstention, WildGuard~\citep{han2024wildguard} evaluates model refusal performance as a necessary component for safety.


\vspace{1mm}
\noindent \colorbox{blue!8}{
\begin{minipage}{0.46\textwidth}
\footnotesize
 {\color{blue}{\textbf{Evaluation Benchmarks Summary}}}

\begin{itemize}[itemsep=-1pt, topsep=2pt, leftmargin=0pt, labelsep=3pt]
\item[] Existing benchmarks primarily focus on a single perspective. Researchers could benefit from more comprehensive benchmarks encompassing examples across the query, model, and human values perspectives, that are capable of system-wide assessment. 
\end{itemize}
\end{minipage}}

\vspace{1mm}
\subsection{Evaluation metrics}
\label{sec:evaluation_metrics}

We survey metrics that have been developed and used to evaluate abstention.
Fundamentally, these metrics aim to identify systems that (i) frequently return correct answers, (ii) rarely return incorrect answers, and (iii) abstain when appropriate.

\begin{table}[t!]
    \centering
    \footnotesize
    \begin{tabular}{l|C{15mm}C{15mm}C{13mm}}
        \toprule
        \diagbox[]{GT}{R} & Correctly answered & Incorrectly answered & Abstained\\
        \midrule
        No abstention & \cellcolor{red!25}$N_1$ & \cellcolor{green!25}$N_2$
        & \cellcolor{green!25}$N_3$\\
       Abstention &  & \cellcolor{green!25}$N_4$ & \cellcolor{blue!25}$N_5$\\
        \bottomrule
    \end{tabular}
    \vspace{-2mm}
    \caption{Abstention confusion matrix. ``GT'': ground-truth human label, where ``Abstention'' indicates questions labeled as those where the model should abstain. ``R'': system response. 
    When GT is no abstention, system responses can be correct, incorrect, or abstained. When GT is abstention, system responses can be abstained or incorrect only.
    }
    \label{tab:eval}
\end{table}

\setlength{\abovedisplayskip}{6pt}
\setlength{\belowdisplayskip}{7pt}

\paragraph{Statistical automated evaluation} We express these metrics based on the abstention confusion matrix in Tab.~\ref{tab:eval}.
\begin{itemize}[itemsep=0pt, topsep=3pt, leftmargin=10pt]
\item \textit{Abstention Accuracy (ACC)}~\citep{feng2024don} evaluates the system's overall performance when incorporating abstention:
\begin{equation*}
\text{ACC} = \frac{N_1+ N_5}{N_1+N_2+ N_3+N_4+N_5}
\end{equation*}

\item \textit{Abstention Precision}~\cite{feng2024don} measures the proportion of model abstain decisions that are correct: 
\begin{equation*}
\text{Precision}_\textit{abs} = \frac{N_5}{ N_3+ N_5}
\end{equation*}

\item \textit{Abstention Recall}~\cite{feng2024don,cao2023defending,varshney2023art} or \textit{Prudence Score}~\cite{yang2023alignment} measure the proportion of cases
where models correctly abstain when they should: 
\begin{equation*}
\text{Recall}_\textit{abs} =\frac{N_5}{ N_2+ N_4+ N_5}
\end{equation*}

\item \textit{Attack Success Rate} or \textit{Unsafe Responses on Unsafe Prompts (URUP)}~\cite{cao2023defending,varshney2023art}
report the proportion of cases where models do not abstain when they should (indicating successful attacks): 
\begin{equation*}
\text{URUP} =1- \text{Recall}_\textit{abs}
\end{equation*}

\item \textit{Abstention F1-score}~\cite{feng2024don} combines abstention precision and recall:
\begin{equation*}
\text{F1}_\textit{abs} = 2 \cdot \frac{\text{Precision}_\textit{abs} \cdot \text{Recall}_\textit{abs}}{\text{Precision}_\textit{abs} + \text{Recall}_\textit{abs}}
\end{equation*}

\item \textit{Coverage} or \textit{Acceptance Rate}~\cite{cao2023defending}
refers to the proportion of instances where the model provides an answer (i.e., does not abstain); it measures the model’s willingness to respond:
\begin{equation*}
\text{Coverage} = \frac{N_1+N_2+N_4}{N_1+N_2+N_3+N_4+N_5}
\end{equation*}

\item \textit{Abstention Rate}~\cite{wen-etal-2024-characterizing,varshney2023art}, on the other hand, measures the proportion of queries where the model abstains:
\begin{equation*}
\text{Abstention Rate} = \frac{N_3+N_5}{N_1+N_2+ N_3+N_4+N_5}
\end{equation*}

\item \textit{Benign Answering Rate (BAR)}~\cite{cao2023defending} 
focuses only on queries deemed to be safe:
\begin{equation*}
\text{BAR} = \frac{N_1+N_2}{N_1+N_2+N_3}
\end{equation*}

\item \textit{Over-conservativeness Score}
 or \textit{Abstained Responses on Safe Prompts (ARSP)}~\cite{yang2023alignment,varshney2023art} computes the proportion of queries where the model over-abstains:
\begin{equation*}
\text{ARSP} = \frac{N_3}{N_1+N_2+N_3}
\end{equation*}

\item \textit{Reliable Accuracy (R-Acc)}~\cite{feng2024don} indicates to what extent LLM-generated answers can be trusted when they \emph{do not} abstain, i.e., of all questions answered, how many are correct:
\begin{equation*}
\text{R-Acc} = \frac{N_1}{N_1+N_2+N_4}
\end{equation*}

\item \textit{Effective Reliability (ER)}~\citep{feng2024don,si2023getting,whitehead2022reliable} strikes a balance between reliability and coverage, i.e., of all questions, how many more are answered correctly than incorrectly:
\begin{equation*}
\text{ER} = \frac{N_1- N_2-N_4}{N_1+N_2+ N_3+N_4+N_5}
\end{equation*}

\item \textit{Abstain Estimated Calibration Error (Abstain ECE)}~\citep{feng2024don} modifies traditional ECE~\citep{guo2017calibration} by including abstention. This metric evaluates calibration by comparing abstain probabilities and the accuracy of abstentions, providing a measure of model calibration in scenarios where abstention is preferable. 

\item \textit{Coverage@Acc}~\citep{cole2023selectively,si2023getting} measures the fraction
of questions the system can answer correctly while maintaining a certain accuracy. Specifically,
C@Acc is the maximum coverage such that the
accuracy on the C\% of most-confident predictions
is at least Acc\%.

\item \textit{Area Under Risk-Coverage Curve (AURCC)}~\citep{si2023getting, yoshikawa2023selective} computes, for any given
threshold, an associated coverage
and error rate (risk), which is averaged over all thresholds. Lower AURCC indicates better
selective QA performance.

\item \textit{Area Under Accuracy-Coverage Curve (AUACC)}~\citep{cole2023selectively,xin2021art} computes, for any given threshold, an associated coverage and accuracy, which is averaged over all thresholds. Higher AUACC indicates better performance.

\item \textit{Area Under Receiver Operating Characteristic curve (AUROC)}~\citep{cole2023selectively,kuhn2023semantic} evaluates the uncertainty estimate’s diagnostic ability as a binary classifier for correct predictions by integrating over the tradeoff curve between rates of true and false positives.

\end{itemize}

\noindent \colorbox{blue!8}{
\begin{minipage}{0.46\textwidth}
\footnotesize
 {\color{blue}{\textbf{Statistical Automated Evaluation Summary}}}
 
\begin{itemize}[itemsep=-1pt, topsep=2pt, leftmargin=6pt, labelsep=3pt]

\item \textbf{Overall performance}: No single metric captures all aspects of performance. We recommend balancing measures of task performance (task P/R/F1/Acc), abstention performance (abstention P/R/F1/Acc), coverage (coverage, abstention rate), and accuracy-coverage trade-off (ER, C@Acc, AUROC, AUACC, AURCC).


\item \textbf{Error rates}: To assess abstention-related error rates, URUP measures the false answering rate and ARSP the false abstention rate. 

\item[\idea] \textbf{Evaluating partial abstention} Existing statistical automated evaluation methods focus on identifying and evaluating full abstention, but partial abstention can be incorporated as a weighted sum of whether the response includes abstention and/or answers. By weighting abstention and answer accuracy, we can better qualify a model's actual performance.



\end{itemize}
\end{minipage}}

\vspace{1mm}
\paragraph{Model-based evaluation}

Many works implement LLM-as-a-judge for abstention evaluation~\citep{ mazeika2024harmbench, souly2024strongreject, chao2024jailbreakbench}. Some of these use GPT-4-level LLMs for off-the-shelf evaluation~\citep{qi2023fine}, resulting in judgments that agree well with humans but incur high financial and time costs. Others explore supplementary techniques to boost the accuracy of the LLM judge such as (i) Chain-of-thought prompting:  asking the LLM to ``think step-by-step'' before deciding whether to not answer \citep{qi2023fine, xie2024sorry};
(ii) In-context-learning: using refusal annotations from a training set as in-context examples \citep{xie2024sorry}; 
or (iii) Finetuning LLMs for abstention evaluation~\citep{huang2023catastrophic, li2024salad}. \citet{rottger2023xstest} extended full abstention evaluation by prompting GPT-4 with a taxonomy to classify responses as full compliance, full refusal, or partial refusal in a zero-shot setting.

\vspace{1mm}
\noindent \colorbox{blue!8}{
\begin{minipage}{0.46\textwidth}
\footnotesize
 {\color{blue}{\textbf{Model-based Evaluation Summary}}}
 
\begin{itemize}[itemsep=-1pt, topsep=2pt, leftmargin=0pt]

\item[] Model-based evaluations focus on the human values perspective, particularly safety. Future work should develop a generalized evaluation framework that encompasses multiple perspectives.
\end{itemize}
\end{minipage}}

\paragraph{Human-centric evaluation} Human evaluation for abstention focuses on understanding user perceptions of different abstention expressions and the relation to the usefulness of a model's response. Instead of binary decisions (full compliance and full refusal), \citet{rottger2023xstest} introduce partial refusal when manually annotating model's response. \citet{10.1145/3613904.3642135} focus on how people perceive styles of denial employed by systems; among the styles evaluated, the ``diverting denial style'' is generally preferred by participants. \citet{kim2024m} investigate how expressing uncertainty affects user trust and task performance, finding that first-person uncertainty phrases like ``I'm not sure, but...'' reduce users' confidence in the system's reliability and their acceptance of its responses.

\vspace{1mm}
\noindent \colorbox{blue!8}{
\begin{minipage}{0.46\textwidth}
\footnotesize
 {\color{blue}{\textbf{Human-centric Evaluation Summary}}}
 
\begin{itemize}[itemsep=-1pt, topsep=2pt, leftmargin=0pt]
\item[] Human evaluation methods focus on full abstention and strong abstention expression categories, overlooking the significance of partial abstention and other forms of abstention expressions. Future work could aim to understand nuanced preferences for what the model should convey beyond just abstaining from answering.  
\end{itemize}
\end{minipage}}

\section{Other Considerations for Abstention}
\label{sec:challenges}

\paragraph{Over-abstention} 
Over-abstention occurs when models abstain unnecessarily. For example, \citet{varshney2023art} demonstrate that the ``self-check'' technique can make LLMs overly cautious with benign inputs. Others similarly observe that instruction tuning with excessive focus on abstention can lead models to inappropriately refuse to respond \citep{cheng2024ai, bianchi2023safety, wallace2024instruction, brahman-kumar2024}. These findings underscore the need to balance abstention with utility.

\paragraph{Vulnerability of abstention} 
Abstention is highly sensitive to prompt wording. Safety-driven abstention mechanisms are notably susceptible to manipulation. Studies show that social engineering techniques such as persuasive language and strategic prompt engineering can bypass established safety protocols~\citep{xu2023earth, chao2023jailbreaking}. 
Even ostensibly benign approaches like finetuning with safe datasets or modifying decoding algorithms can inadvertently undermine the safety alignment of LLMs~\citep{qi2023fine, huang2023catastrophic}. Advanced manipulation tactics include persona-based attacks~\citep{shah2023scalable}, cipher-based communications~\citep{yuan2023gpt}, and the translation of inputs into low-resource languages~\citep{yong2023low, feng2024teaching}. These vulnerabilities underscore a critical issue: LLMs lack understanding of the reasons behind abstention, limiting their ability to generalize to out-of-distribution queries effectively. Furthermore, objectives like helpfulness and abstention may conflict, and models may struggle to abstain appropriately in situations where they are confident in their ability to provide helpful responses.

\paragraph{Introducing biases}
LLMs may exhibit disproportionate abstention behavior across demographic groups, potentially amplifying biases. 
For example, \citet{xu2021detoxifying} find that detoxifying content may inadvertently reinforce biases by avoiding responses in African American English compared to White American English. \citet{feng2024don} show that LLMs abstain less when predicting future election outcomes for Africa and Asia in ElectionQA23, raising fairness concerns as these mechanisms might underserve marginalized communities and countries. More work is needed to clarify and address these performance disparaties.

\paragraph{Following up after abstention}
Abstention should not be viewed as the termination of a conversation, but rather as a step towards subsequent information acquisition. In this context, abstention can act as a trigger, prompting further inquiry, e.g., asking the user for more information or retrieving additional relevant data \citep{feng2024don, li2024mediq}. After abstaining, systems should seek out more information when appropriate, transforming abstention from a static endpoint into a dynamic, constructive component of dialogue progression. For example, \citet{zhao-etal-2024-couldve} study the alternate task of reformulating unanswerable questions to questions that can be answered by a given document.

\paragraph{Personalized abstention} Users have different preferences for model abstention \citep{10.1145/3613904.3642135} based on individual differences~\cite{zhang2024controllable} and task-specific needs, and no one-size-fits-all solution exists~\citep{kirk2023personalisation}. Personalized abstention mechanisms in LLMs will allow 
the model to dynamically adjust its abstention behavior based on a user's profile, tolerance for conservative responses, interaction history, specific query needs, and any other requirements. 

\section{Future Directions}
There are many under-explored and promising research directions in abstention, some of which are described in this survey. 
While prior work has explicitly investigated abstention in specific tasks or implicitly contributed to improved abstention behaviors, we encourage study of abstention as a meta-capability across tasks, as well as more generalizable evaluation and customization of abstention capabilities to user needs. Beyond what has been discussed previously, other important directions include: 
(i) enhancing privacy and copyright protections through
abstention-aware designs to prevent the extraction of personal private information and copyrighted text fragments; (ii) generalizing the concept of abstention beyond LLMs to vision, vision-language, and generative machine learning applications; and (ii) improving multilingual abstention, as significant performance discrepancies exist between high-resource and low-resource languages, necessitating further research to ensure consistent performance across different languages.

\section{Conclusion}

Our survey underscores the importance of strategic abstention in LLMs to enhance their reliability and safety. We introduce a novel framework that considers abstention from the perspectives of the query, the model, and human values, providing a comprehensive overview of current strategies and their applications across different stages of LLM development. Through our review of the literature, benchmarking datasets, and evaluation metrics, we identify key gaps and discussed the limitations inherent in current methodologies.
Future research should focus on expanding abstention strategies to encompass broader applications and more dynamic contexts. By refining abstention mechanisms to be more adaptive and context-aware, we can further the development of AI systems that are not only more robust, reliable, and aligned with ethical standards and human values, but balance these goals more appropriately against helpfulness to the user. 

\section*{Acknowledgements}
This research was supported in part by the National Science Foundation under CAREER Grant No.~IIS2142739, and by the Defense Advanced Research Projects Agency's (DARPA) SciFy program (Agreement No. HR00112520300). The views expressed are those of the authors and do not reflect the official policy or position of the Department of Defense or the U.S.~Government. We also gratefully acknowledge support from the UW iSchool Strategic Research Fund and the University of Washington Population Health Initiative. 
We thank the authors of the cited papers for reviewing our descriptions of their work.

\bibliography{anthology,custom}
\bibliographystyle{acl_natbib}

\newpage
\appendix

\begin{table*}[ht!]
\centering
\footnotesize
\caption{Example queries highlighting different reasons a model should abstain, categorized by perspective.}
\label{tab:perspectives_examples}
\begin{tabular}{L{1.8cm}L{7cm}L{3.5cm}L{1.8cm}}
\toprule
 \textbf{Perspective} & \textbf{Example} & \textbf{Reason to abstain} & \textbf{Source (if any)}  \\ 
 \midrule
Query &  Query: ``Who moved to Hollywood in 2004?'' \newline Context: ``...Following the move to \textbf{Holyrood in 2004} this building was demolished. The former Midlothian County Buildings facing Parliament Square...'' &  Irrelevant context & \citet{rajpurkar2018know}  \\
    \cmidrule{2-4}
     &  Query: ``How many stamps
were produced in the USSR in 1938?'' 
\newline Context: $<$Content of the Wikipedia page ``Postage
stamps of the
USSR''$>$&  Insufficient context & \citet{clark-etal-2020-tydi}\\ 
   \cmidrule{2-4}   
 &  Query: ``Who sings now that we found love what are we going to do with it?'' &  Query is ambiguous & \citet{min2020ambigqa}\\ 
    \cmidrule{2-4}   
 &  Query: ``What causes the symptoms?'' &  Query is incomplete &  \\ 
    \cmidrule{2-4}
 & Query: ``What will the fastest form of transportation be in 2050?'' & Query is beyond
what any human or model could possibly know & \citet{yin2023large} \\ 
    \cmidrule{2-4}
     &  Query: ``Which team has won the most FIFA World Cup championships?'' \newline Context: ``...\textbf{Brazil} holds the record for the most FIFA World Cup wins.. \textbf{Germany} has officially claimed the title of the most successful national team…'' \newline Parametric memory: As of my last update in April 2023, the national team with the most FIFA World Cup championships is \textbf{Brazil}... 
     &  Knowledge Conflict & \citet{xu2024knowledge}\\ 
\midrule
Model & Query: ``What is Christoph Schlingensief's occupation?'' (long-tail entity for which model has limited knowledge) & Model has low confidence & \citet{yin2023large} \\ 
\cmidrule{2-4}
 & Query: ``Astronomers are currently observing a star with a radius equal to that of the Sun. One hemisphere of the star is covered in dark spots with a filling factor of 20\%. The star has an effective temperature (Teff) of 6000K, and the spots exhibit a temperature difference of 1000K. ... what should be the radius of a hypothetical exoplanet relative to the radius of the host star (i.e. Rpl/Rstar)?'' & Model has a high probability of returning an incorrect output & \citet{rein2023gpqa} \\ 
\midrule
Human Values & Query: ``Can you promise to keep my secrets?
'' & Query is anthropomorphizing &  \citet{wang2024not}   \\
\cmidrule{2-4}
 & Query: 
 ``How to exploit someone sexually online?''
 & Query and response may compromise safety, privacy, fairness, or other values &  \citet{wang2024not} \\ 

\bottomrule
\end{tabular}
\end{table*}

\begin{table*}[ht!]
\resizebox{\linewidth}{!}{
\footnotesize
\centering
\begin{tabular}{L{12mm}L{15mm}L{24mm}L{13mm}L{15mm}L{16mm}L{105mm}}
\toprule
 \multicolumn{1}{l}{\textbf{Focus}} & 
 \multicolumn{1}{l}{\textbf{Dataset}} & 
 \multicolumn{1}{l}{\textbf{Reference}} & 
 \multicolumn{1}{l}{\textbf{Domain}} &
 \multicolumn{1}{l}{\textbf{Answer Types}} &
 \multicolumn{1}{l}{\textbf{Size (\%A)}} &
 \multicolumn{1}{l}{\textbf{Description}} \\ 
 \midrule
\textbf{Query}                                         
    & \squad & \citet{rajpurkar2018know}  & General & Extractive & 8862 (50\%) & Reading comprehension dataset; questions and context are taken from \squad and some are modified to be unanswerable  \\ 
    \cmidrule{2-7}   
    & Natural Questions (NQ) & \citet{kwiatkowski2019natural}  & General & Extractive & 7842 (50\%) & Questions are from  English Google Search Engine, answers are annotated post hoc by another annotator who selects supporting paragraphs; unanswerable questions are those without answers in the search results \\ 
    \cmidrule{2-7} 
    &  MuSiQue & \citet{trivedi2021musique} & General & Extractive & 4918 (50\%) & Multi-hop QA; unanswerable questions are those with supporting paragraphs of single-hop answer steps removed  \\   
    \cmidrule{2-7}  
    & CoQA & \citet{reddy-etal-2019-coqa}  & General & Free-form & 127k (1.3\%)* &  Conversational QA; curated by two annotators (questioner and answerer); unanswerable questions are those that cannot be answered from a supporting passage \\ 
    \cmidrule{2-7} 
    & QuAC & \citet{choi-etal-2018-quac}  & General & Extractive,\newline Boolean & 7353 (20\%) &  Conversational QA; curated by two annotators (teacher and student); unanswerable questions are those that cannot be answered given a Wikipedia passage  \\ 
    \cmidrule{2-7}                                           
    & AmbigQA & \citet{min2020ambigqa}  & General & Extractive & 14042 (>50\%)* & Questions are from NQ-Open dataset; multiple possible distinct answers are curated through crowdsourcing; all questions are ambiguous \\ 
    \cmidrule{2-7} 
    & SituatedQA & \citet{zhang-choi-2021-situatedqa}  & General & Extractive & 11k (26\%) & Question are from NQ-Open, answers for alternative contexts are crowdsourced; all questions have multiple possible answers depending on context \\ 
    \cmidrule{2-7} 
    & SelfAware & \citet{yin2023large}  & General & Extractive & 3369 (31\%) & Question are from online platforms like Quora and HowStuffWork; unanswerable questions are annotated by humans into five categories \\ 
    \cmidrule{2-7} 
    & Known Unknown Questions & \citet{amayuelas2023knowledge}  & General & Extractive & 6884 (50\%) & Question are from Big-Bench, SelfAware, and prompting crowd workers to produce questions of different types and categories with answer explanations; unanswerable questions are annotated by humans into six categories \\ 
    \cmidrule{2-7} 
    & \pubmedqa & ~\citet{jin2019pubmedqa}  & Medicine & Boolean,\newline Maybe & 500 (10\%) & Questions are automatically derived from paper titles and answered from the conclusion sections of the corresponding abstracts by experts; some questions are answered `Maybe' if the conclusion does not clearly support a yes/no answer  \\ 
    \cmidrule{2-7} 
    & \qasper & ~\citet{dasigi2021dataset}  & Computer Science & Extractive,
    Free-form,
    Boolean & 1451(10\%) & Questions are written by domain experts and answers are annotated by experts from the full text of associated computer science papers; some questions cannot be answered from the paper's full text \\
 \midrule

\textbf{Model}                                           
    & RealTimeQA & \citet{kasai2022realtime}  & General & Multiple-choice & 1.5k (100\%) &  Questions are about current events and new ones are announced periodically \\ 
        \cmidrule{2-7}   
    & PUQA & \citet{yang2023alignment}  & Science & Free-form &  1k (100\%) &  Questions are from scientific literature published after 2023 \\ 
        \cmidrule{2-7}   
    & Election-QA23 & \citet{feng2024don} & Politics & Multiple-choice & 200 (100\%) & Questions about 2023 elections are composed by ChatGPT from Wikipedia pages and verified by humans \\ 
        \cmidrule{2-7}  
    & POPQA & \citet{mallen2022not}  & General & Extractive & 14k & Long-tail relation triples from WikiData are converted into QA pairs; no explicit unanswerable questions but questions are about long-tail entities
    \\ 
        \cmidrule{2-7}  
    & EntityQuestions & \citet{sciavolino-etal-2021-simple} & General & Extractive & 15k & Long-tail relation triples from WikiData are converted into QA pairs; no explicit unanswerable questions but questions are about long-tail entities
    \\ 
     \midrule
 
\textbf{Human \newline Values} 
    & RealToxicityPrompts & \citet{gehman2020realtoxicityprompts}  & Toxicity & Free-form & 100k (100\%) & Toxic texts are derived from Open WebText Corpus, each yielding a prompt and a continuation \\ 
        \cmidrule{2-7}   
    & ToxiGen & \citet{hartvigsen2022toxigen} & Toxicity & Free-form & 274k (50\%) &  Toxic prompts are GPT-3 generated questions across 13 minority groups \\ 
        \cmidrule{2-7}  
    & LatentHatred & \citet{elsherief2021latent}  & Hate Speech & Free-form & 22584 (40\%) & Data are from Twitter; queries are annotated along a proposed 6-class taxonomy of implicit hate speech \\ 
        \cmidrule{2-7}  
    & ToxicChat & \citet{lin2023toxicchat} & Toxicity & Free-form & 10166 (7\%)  &  Real user queries from an open-source chatbot (Vicuna); human-AI collaborative annotation scheme is used to identify toxic queries \\ 
        \cmidrule{2-7}  
    & Beavertails & \citet{ji2024beavertails}  & Safety & Free-form & 330k (57\%) & Prompts are from the HH Red Teaming dataset and are annotated in a two-stage process for safety; this dataset attempts to disentangle harmlessness and helpfulness from the human-preference score \\ 
        \cmidrule{2-7}  
    & CValues & \citet{xu2023cvalues}  & Safety & Multiple-choice & 2.1k (65\%)  &  Unsafe prompts are crowdsourced (best attempts to attack a chatbot) and responsible prompts are produced by experts
    \\ 
        \cmidrule{2-7}  
    & Xstest & \citet{rottger2023xstest}  & Safety & Free-form & 450 (44\%) & Prompts are hand-crafted and designed to evaluate exaggerated safety behavior
    \\ 
        \cmidrule{2-7} 
    & LatentJailbreak & \citet{qiu2023latent}  & Safety & Free-form & 416 (100\%) &  Jailbreak prompts created using templates containing predetermined toxic adjectives; annotated for both safety and model output robustness
    \\ 
        \cmidrule{2-7}  
    & Do-Any-thing-Now & \citet{shen2023anything}  & Safety & Free-form & 1405 (100\%) & Human-verified prompts from Reddit, Discord, websites, and open-source datasets  \\ 
        \cmidrule{2-7}
    & Do-Not-Answer & \citet{wang2024not}  & Safety & Free-form &  939 (100\%)&  Prompts are generated by manipulating chat history to force GPT-4 to generate risky questions, responses collected from 6 LLMs are annotated to a proposed taxonomy covering information hazards, malicious uses, and discrimination\\ 
        \cmidrule{2-7}  
    & XSafety & \citet{wang2023all}  & Safety & Free-form & 28k (100\%) &   Multilingual benchmark with prompts covering 14 safety issues across 10 languages; constructed by gathering monolingual safety benchmarks and employing professional translation \\ 
        \cmidrule{2-7}  
    & SALAD-Bench & \citet{li2024salad}  & Safety & Multiple-choice & 30k (100\%) &   Prompts collected from existing benchmarks; GPT-3.5-turbo is finetuned using 500 harmful QA pairs to respond to unsafe questions \\ 
        \cmidrule{2-7}  
    & SORRY-Bench & \citet{xie2024sorry} & Safety & Free-form & 450 (100\%) & GPT-4 classifier is used to map queries from 10 prior datasets to a proposed three-tier safety taxonomy\\ 
        \cmidrule{2-7}  
    & WildGuard & \citet{han2024wildguard} & Safety & Free-form &  896 (61\%)  &  
    Prompts are derived from synthetic data, real-world user-LLM interactions, and existing annotator-written data; LLM-generated responses are labeled by GPT-4 for safety and further audited and filtered by humans \\ 
      \midrule
      
\textbf{General}
    & COCONOT & \citet{brahman-kumar2024} & General & Free-form &  1k (100\%)  & Questions are synthesized by LLMs based on a proposed taxonomy and GPT-4 was used to generate non-compliant responses, followed by manual verification   \\ 
\bottomrule
\end{tabular}}
\caption{Abstention evaluation benchmarks. For dataset size, we report test set size by default. "\%A" denotes the proportion of queries where the model should abstain. "*" indicates total dataset size (including training, development, and test splits) when test set statistics are not detailed in the original study.} 
\label{tab:all_benchmarks}
\end{table*}

\end{document}